\pdfoutput=1
\documentclass{article}

\usepackage[nonatbib, final]{nips_2016}

\usepackage{array, makecell} %

\usepackage[utf8]{inputenc}
\usepackage{amsmath}
\usepackage{amssymb}
\usepackage[T1]{fontenc}    % use 8-bit T1 fonts
\usepackage{url}            % simple URL typesetting
\usepackage{booktabs}       % professional-quality tables
\usepackage{amsfonts}       % blackboard math symbols
\usepackage{nicefrac}       % compact symbols for 1/2, etc.
\usepackage{microtype}      % microtypography
\usepackage{tikz}
\usepackage{caption}
\usetikzlibrary{bayesnet}
\usepackage{float}
\usepackage{graphicx}
\usepackage{color}

\begin{document}

\title{\Large{A deep generative model for gene expression profiles from single-cell RNA sequencing}}

\author{
  Romain Lopez$^\dagger$, Jeffrey Regier$^\dagger$, Michael Cole$^{\dagger\dagger}$, Michael Jordan$^\dagger$ and Nir Yosef$^\dagger$\\
    $\dagger$~Department of Electrical Engineering and Computer Sciences, University of California, Berkeley \\
    $^{\dagger\dagger}$~Department of Physics, University of California, Berkeley \\
      \texttt{\{romain\_lopez, regier, mbcole, jordan, niryosef\}@berkeley.edu} \\
}
\setlength{\abovedisplayskip}{5pt}
\setlength{\belowdisplayskip}{5pt}
\setlength{\abovedisplayshortskip}{5pt}
\setlength{\belowdisplayshortskip}{5pt}

\maketitle

%Single-cell RNA expression analysis (scRNA-seq) is revolutionizing whole-organism science [1, 2] allowing the unbiased identification of previously uncharacterized molecular heterogeneity at the cellular level. Statistical analysis of single-cell gene expression profiles can highlight putative cellular subtypes, %delineating subgroups of T cells [3], lung cells [4] and myoblasts [5]. These subgroups can be clinically relevant: for example, individual brain tumors contain cells from multiple types of brain cancers, and greater tumor heterogeneity is associated with worse prognosis [6].

%Despite the success of early single-cell studies, the statistical tools that have been applied to date are largely generic, rarely taking into account the particular structural features of single-cell expression data. In particular, single-cell gene expression data contain an abundance of dropout events that lead to %zero expression measurements. These dropout events may be the result of technical sampling effects (due to low transcript numbers) or real biology arising from stochastic transcriptional activity (Fig. 1 a).

\vspace{-0.4 cm}

%Gene expression profiling through the lens of single-cell RNA sequencing promises to reveal the secrets of cellular identity and molecular circuitry~\cite{Wagner2016}. After capturing mRNA transcripts at the cell-level across a whole tissue, alignment and estimation methods produce a data matrix 
%Gene expression levels are key to understanding many aspects of cells function.
%Single-cell RNA sequencing, a new technology, allows investigators to measure snapshots of transcription in individual cells. RNA-Seq alignment and expression estimation methods produce a data matrix 
Single-cell RNA sequencing (scRNA-Seq) is a revolutionary technology, which allows studying fundamental biological questions that were previously out of reach \cite{Wagner2016, Tanay2017}. It allows, for the first time, to reveal a cell's identity and characterize its molecular circuitry in an unbiased, data-driven way. The product of a scRNA-Seq experiment is a data matrix $X$ where entry $X_{ng}$ approximates the number of transcripts of gene $g$ in cell $n$. Careful computational analysis allows deriving from such data exciting insights in diverse biomedical fields \cite{Zeng2017, Stubbington2017}. 
While it is typical to observe thousands of gene products per cell, many
transcripts are observed very infrequently, and for technical reasons related to
the method of sequencing these are particularly prone to high variance.
Additionally, due to the limited transcript capture efficiency inherent to
RNA-Seq protocols, entries of $X$ are typically zero-inflated~\cite{ZIFA}.

While there is often little prior knowledge of single-cell heterogeneity
generating $X$, a reasonably general assumption is that $X$ has been generated
from a low-dimensional manifold of cellular states~\cite{Wagner2016}. Therefore,
numerous dimensionality reduction techniques have been proposed for interpreting
$X$ (e.g., to facilitate clustering, visualization, and data imputation). Each
technique has shortcomings, however. Most are based on linear models of the data
\cite{ZIFA,ZINB-WAVE,Detomaso2016} though there is no basis for assuming
linearity. Most are optimized with batch algorithms, preventing them from
scaling beyond thousands of cells \cite{ZIFA,ZINB-WAVE,BISCUIT}. However,
sequencing millions of cells is becoming possible~\cite{10x}. The best
performing method to date \cite{ZINB-WAVE} is particularly complicated to train,
involving numerous subroutines for alternating minimization. Recent articles
apply neural networks, but without an architecture based on
biology~\cite{Ding2017, Lin2017}.

We propose Single-cell Variational Inference (scVI), a probabilistic inference
procedure based on a fully generative model.
In scVI, some conditional distributions are specified by neural networks
that encode complex, nonlinear relationships, learned from large datasets.
scVI explicitly models technical effects, so as to ``disentangle'' them from
a low-dimensional vector that represents the cells' underlying states.

After describe our generative model (Section~\ref{model})
and an inference procedure for it (Section~\ref{inference}),
we compare scVI to alternative methods on three benchmarks tasks (Section~\ref{benchmarks}).
Finally, we introduce a Bayesian hypothesis testing procedure that leverages scVI to find
differentially expressed genes (Section 4). Our source code, based on TensorFlow, is publicly
available at \url{https://github.com/YosefLab/scVI}.

\vspace{-0.2cm}

\section{The scVI probabilistic model} \label{model}
Figure \ref{graph} represents the probabilistic model graphically.
Latent variable
\begin{align*}
z_n  \sim \mathcal N(0, I)
\end{align*}
is a low-dimensional random vector describing cell $n$. For neural
network $f_w$, latent variable
\begin{align*}
w_{ng} \sim \mathrm{Gamma}(f_w(z_n, \gamma_n))
\end{align*}
accounts for the stochasticity of gene $g$ expressed in cell $n$.
Here the constant $\gamma_n$ are optional covariates that can be passed to $f_w$ that account for batch effects, like normalization~\cite{Vallejos2017,Risso2014},
to remove unwanted variation from the latent representation.
Latent variable
\begin{align*}
y_{ng} \sim \mathrm{Poisson}(w_{ng})
\end{align*}
is the underlying expression level for gene $g$ in cell $n$. 

\begin{minipage}{0.75\textwidth}
For neural network
$f_h$, latent variable
\begin{align*}
h_{ng} \sim \mathrm{Bernoulli}(f_h(z_n, \gamma_n))
\end{align*}
indicates whether a particular entry has been
“zeroed out” due to technical effects~\cite{ZIFA,ZINB-WAVE}.

Finally, the observed gene expression level is defined by:
\begin{align*}
x_{ng} =
\begin{cases}
y_{ng} & \text{ if } h_{ng} = 0,\\
0 & \text{ otherwise}.\\
\end{cases}
\end{align*}

Conditional distribution $p(x_{ng} | z_{n})$ is a zero-inflated
negative binomial---a distribution known to effectively model the kinetics of
stochastic gene expression with some entries replaced by zeros~\cite{Grun2014}.

The neural networks $f_w$ and $f_h$ use dropout regularization and batch
normalization. Each network has 3 fully connected-layers, with 128 nodes each.
The activation functions are all ReLU, exponential, or linear.
Weights for some layers are shared between $f_w$ and $f_h$.

\end{minipage}\hfill
\begin{minipage}{0.20\textwidth}
  \centering
  \tikz{ %
  \node[obs] (x) {${x}_{ng}$} ; %
  \node[latent, above=0.7 of x, xshift=-0.8cm] (h) {${h}_{ng}$};
  \node[latent, above=0.2 of x, xshift=0.8cm]  (y) {${y}_{ng}$};
  \node[latent, above=0.4 of y] (w) {${w}_{ng}$};
  \node[latent, above=0.4 of w, xshift=-0.8cm] (z) {${z}_{n}$};

    %\node[latent, right=of theta] (z) {z} ; %
    %\node[latent, above=of z] (beta) {$\beta$} ; %
    \plate[inner sep=0.15cm, xshift=-0.cm, yshift=0.cm] {plate1} {(w) (h) (y) (x)} {G}; %
   \plate[inner sep=0.15cm, xshift=-0.cm, yshift=0.cm] {plate2} {(z) (w) (h) (y) (x) (plate1)} {N}; %
    \edge {z} {w, h} ; %
    \edge {h} {x} ; %
    \edge {w} {y}; %
    \edge {y} {x} ; %
  }

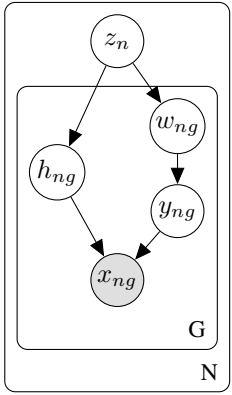
\captionof{figure}{The scVI graphical model}
\label{graph}
\end{minipage}

\section{Posterior inference} \label{inference}
The posterior distribution combines the prior knowledge with information
acquired from the data $X$. We cannot directly apply Bayes rule to determine the
posterior because the denominator (the marginal distribution) $p(x_n)$ is
intractable.

Instead, we use variational inference~\cite{blei2017variational} to approximate
the posterior $p(x_n | z_n)$. Our variational distribution $q(z_n | x_n)$ is
Gaussian with a diagonal covariance matrix. The variational distribution's mean
and covariance are given by an encoder network applied to $x_n$, as
in~\cite{Kingma2013}. The encoder network may, optionally, be given the constant
covariates $\gamma_n$ (along with $x_n$) if we wish to discourage $z_n$ from
encoding batch effects and other unwanted variations.

The variational lower bound is
\begin{align}
\mathcal{L}(n) \geq \mathbb{E}_{q(z|x)}\log p(x|z) - KL(q(z|x)||p(z))
\end{align}

To optimize the variational lower bound, we write $p(x|z)$ by analytically
marginalizing out the discrete random variables $h_{ng}$, $w_{ng}$, and
$y_{ng}$. Now, our variational lower bound is continuous and end-to-end
differentiable. We maximize the variational lower bound using stochastic
backpropagation.

\section{Performance benchmarks} \label{benchmarks}

We assess the performance of scVI at three benchmarks tasks:
generalizing to heldout data  (Section~\ref{heldout}),
imputating ``zeroed out'' data (Section~\ref{impute}), and
recovering known clusters (Section~\ref{clusters}).
Throughout, we compare scVI with factor analysis, as well as two state-of-the-art methods: ZIFA~\cite{ZIFA} and ZINB-WaVE~\cite{ZINB-WAVE}

Only scVI and factor analysis scale to the larger benchmark datasets---a key
advantage relative to ZIFA and ZINB-WaVE.
ZIFA and ZINB-WaVE are based on batch optimization algorithms. Their runtimes for \textit{each} iteration of their numerical optimization routines scale linearly in the number of samples and linearly in the number of genes---both potentially very large. For 10,000 cells, each of these methods requires more than 20 minutes of computation. For 100,000 cells, both methods run out of memory on a machine with 32 GB RAM. scVI trains on the entire 1.3 million cell dataset in less than two hours on a single GPU, using off-the-shelf neural
network software.

\subsection{Generalization to heldout data}
\label{heldout}

For this task, we use a dataset that contains 1.3 million brain cells from \textsc{10x Genomics}~\cite{10x} with 720 sampled variable genes. For each method, we learn a mapping from the 10-dimensional latent space to a reconstruction of training set $X$. Then, we assess the marginal likelihood of held-out data,
conditioned on a latent representation learned for the held-out data by each
model. Table~\ref{LL} shows that scVI best compresses the held-out data,
even for our smallest dataset. scVI's lead over the other methods grows as
the dataset size grows.

\begin{table}
\centering
\begin{tabular}{lccc}
  cells& 4k & 10k & 100k \\
  \hline
  FA & -1178.2 & -1177.3 & -1169.8 \\
  ZIFA & -1250.9 & -1250.7 & NA \\
%  ZINB-VAE& \textbf{-1174.3} & \textbf{-1168.44} & \textbf{-1161.62}\\
%  \hline
  ZINB-WaVE & -1166.3 & -1164.4 & NA\\
  scVI & \textbf{-1159.9} & \textbf{-1147.8} & \textbf{-1128.73}\\
  \hline
\end{tabular}
\vspace{0.2cm}
 \caption{Marginal log likelihood for a held-out subset of the brain cells dataset. NA means we could not run the given algorithm for this sample size.}
 \label{LL}
\end{table}

\subsection{Imputating zeroed-out data}
\label{impute}
On a 10,000 sample of the same dataset, we set to zero entries at random conditioned on the expected transcript abundance --- according to probabilities from the ZIFA model --- to mimic the technical effects that zero out some entries of the real data. Because we have introduced these zeros synthetically, we know 1) each entry's true value, and 2) that each entry is zero because of a
technical effect, not because the true expression level is nearly zero. We also compare for this task to a state-of-the-art method MAGIC~\cite{VanDijk2017} based on diffusion in the cell k-nearest neighbors graph and report results on Table~\ref{IM}.

\begin{table}
\centering
\begin{tabular}{lccc}
  &     imputation        & identification of \\
  &  error  & zeroed-out\\
  \hline
  ZIFA & 3.00 & 1.955 \\
  MAGIC & 1.806 & NA \\
  ZINB-WaVE & 1.053  & 1.366 \\
  scVI & \textbf{1.048} & \textbf{0.742}  \\
  \hline
\end{tabular}
\vspace{0.2cm}
\caption{Absolute errors for imputing zeroed entries (column 1),
mean cross entropy for predicting which entries were zeroed-out entries (column 2). Scores are based on a dataset of 10,000 brain cells. MAGIC does not predict dropout probabilities.}
\label{IM}

\end{table}

\subsection{Recovering known clusters}
\label{clusters}

To further assess the models, we compare how each clusters cells of
known types (e.g., muscle cells, blood cells) in latent space.
For this task, we make a slight modification to our model: we treat each $z_n$ as an unknown parameter to estimate rather than a latent variable with a distribution. This way, our procedure
maximizes mutual information between $z_n$ and $x_n$~\cite{zhao2017infovae}.

A first dataset from~\cite{Zeisel1138} contains 3005 mouse cortex cells and gold-standard labels for seven distinct cell types. Each cell type corresponds
to a cluster to recover. We sample 558 variable genes as in~\cite{BISCUIT}
and report silhouette (a measure of distance between clusters) on the mouse cortex dataset in Table~\ref{SIL}.

\begin{table}
\parbox{.30\linewidth}{
\centering
\begin{tabular}{lc}
  & silhouette \\
  \hline
  FA  & 0.208 \\
  ZIFA  & 0.202 \\
  ZINB-WaVE  & 0.260 \\
%  ZINB-VAE  & 0.245 \\
  scVI  & \textbf{0.285} \\
  \hline
\end{tabular}
\vspace{0.2cm}
\caption{Silhouette scores on the mouse cortex dataset.}
\label{SIL}
}
\hfill
\parbox{.65\linewidth}{
\centering
\begin{tabular}{lcc} 
  & silhouette & QC correlation\\
  & (higher is better)&  (lower is better)\\
  \hline
  PCA & 0.314 & 0.381 \\
  PCA (normalized)  & 0.321 & 0.169 \\
  scVI (no covariates)  & 0.375 & 0.366 \\
  scVI & \textbf{0.379} & \textbf{0.157} \\
  \hline
\end{tabular}
\vspace{0.2cm}
\caption{Unwanted variation metric on the PBMCs dataset. \\ ~ \\ ~}
\label{UV}
}
\end{table}

A second dataset contains 12039 Peripheral blood mononuclear cells (PBMCs) from \cite{Zheng2017} with 10310 sampled genes and get biologically meaningful clusters with the software Seurat~\cite{SEURAT}.
For this dataset, we use SCONE~\cite{SCONE} to select most important factors of unwanted variation to be incorporated into downstream models. These factors generally include batches meta-data, sequencing depth (number of transcript per cell) and quality controls (QC) for each cell. In this case, SCONE selected a strategy that consists in scaling by depth and regressing out the QC.

We compare scVI with and without covariates with a PCA with and without normalization in Table~\ref{UV} and show we can better remove the variation while yielding a high silhouette score --- which means we would get a consistent but tighter clustering with our latent space.

\section{Differential expression}

A significant application of our generative model and of main interest in the field is to go from a clustering to a procedure for identifying gene differentially expressed between two cell-types. Our model relies on Bayesian statistics and can thus benefit from uncertainty evaluation to provide a hypothesis testing framework for differential expression.

Let $A$ and $B$ be two set of cells and $g$ a fixed gene. Now take $(a, b) \in A\times B$ and say we want to test the following:

$$\mathcal{H}_0^g: w_{ag} < w_{bg} \textrm{~~~~vs.~~~~}\mathcal{H}_1^g: w_{ag} \geq w_{bg}$$

 where $w$ is the Gamma latent variable in the generative model, i.e the mean of the gene expression conditioned on a non-dropout event.  The posterior of these hypotheses can be approximated via the variational distribution:

 $$ p(\mathcal{H}_0^g | x) \approx \iint_{z_a, z_b, w_{ag}, w_{bg}} p(w_{ag} < w_{bg})dp(w_{ag} | z_a)dq(z_a | x_a)dp(w_{bg} | z_b)dq(z_b | x_b)$$

 where all the measures are uni-dimensional or low-dimensional so we can use naive monte-carlo to compute these integrals. We can then use a Bayes factor for the test.

We use again the PBMC dataset from \cite{Zheng2017} and the Seurat-based cell classification to understand how differential expression is captured by our testing method compared to tradition DESeq2~\cite{Love2014}. We defined a reference from a publicly-available bulk array expression profiling data for human B cells (n=10) and Dendritic cells (n=10) at baseline of vaccination~\cite{GEO:GSE29618} which we use to test the association of each gene's expression with biological class, defining a 2-sided t-test p-value per gene.

\begin{figure}[h]
\centering
\includegraphics[width=0.6\textwidth]{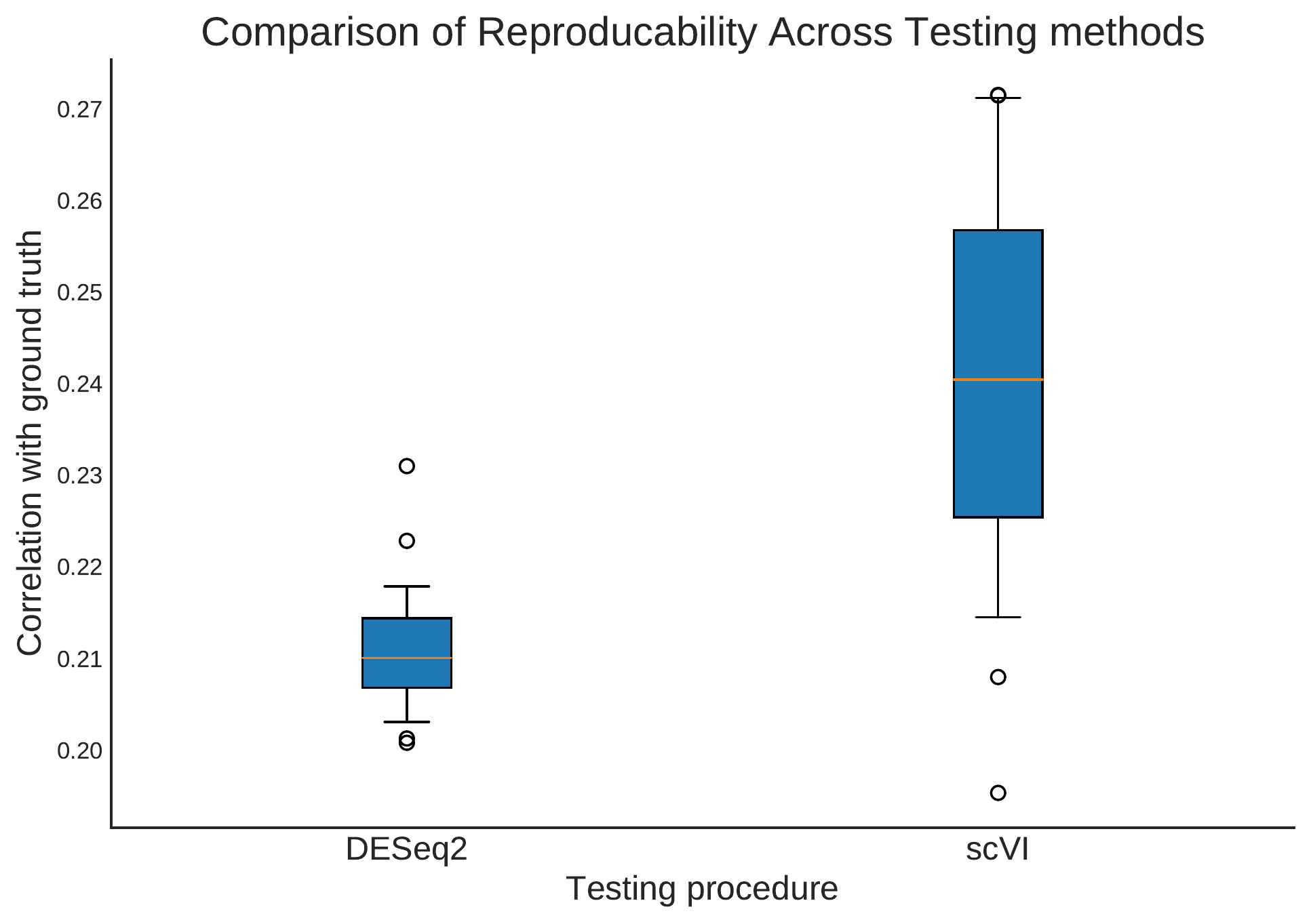}
\caption{Results on the Differential expression task on B cells against DC cells}
\label{de-boxplot}
\end{figure}

Because defining a threshold and then use a ROC curve is ambiguous we prefer to look at reproducibility between the microarray experiment and a family of tests used on the scRNA-Seq sequencing experiment. To quantify this, we model the relationship between significance ranks using the Irreproducible Discovery Rate model for matched rank lists~\cite{Li2011} and report correlation score of the reproducible components in Figure~\ref{de-boxplot}.

\bibliographystyle{unsrt}
\bibliography{biblio}

\end{document}